\pdfoutput=1

\documentclass[11pt]{article}
\usepackage{tcolorbox}
\usepackage[]{acl}

\usepackage{times}
\usepackage{latexsym}
\usepackage{booktabs}
\usepackage{xcolor}
\usepackage{comment}
\usepackage{multirow}
\usepackage[T1]{fontenc}

\usepackage[utf8]{inputenc}

\usepackage{microtype}

\usepackage{inconsolata}

\usepackage{graphicx}

\tcbuselibrary{skins,breakable}
\title{Improving  Automatic Evaluation of Large Language Models (LLMs) in Biomedical Relation Extraction via LLMs-as-the-Judge}

 \author{Md Tahmid Rahman Laskar\thanks{\hspace{0.115cm}Corresponding Emails: \{tahmedge, jhuang\}@yorku.ca}\textsuperscript{,\textdaggerdbl}\textbf{,} 
 \textbf{Israt Jahan}\textsuperscript{\textdaggerdbl}\textbf{,} \\ \textbf{Elham Dolatabadi}\textsuperscript{\textdaggerdbl}\textsuperscript{,\textsection}\textbf{,} 
 \textbf{Chun Peng}\textsuperscript{\textdaggerdbl}\textbf{, }\textbf{Enamul Hoque}\textsuperscript{\textdaggerdbl}\textbf{, }\textbf{Jimmy Xiangji Huang\textsuperscript{\footnotemark[1]}\textsuperscript{,\textdaggerdbl}} \\
            {\textsuperscript{\textdaggerdbl}York University}, 
{\textsuperscript{\textsection}Vector Institute} \\ Toronto, Ontario, Canada         
          \\ }

\begin{document}
\maketitle
\begin{abstract}
Large Language Models (LLMs) have demonstrated impressive performance in biomedical relation extraction, even in zero-shot scenarios. However, evaluating LLMs in this task remains challenging due to their ability to generate human-like text, often producing synonyms or abbreviations of gold-standard answers, making traditional automatic evaluation metrics unreliable. On the other hand, while human evaluation is more reliable, it is costly and time-consuming, making it impractical for real-world applications. This paper investigates the use of LLMs-as-the-Judge as an alternative evaluation method for biomedical relation extraction. We benchmark 8 LLMs as judges to evaluate the responses generated by 5 other LLMs across 3 biomedical relation extraction datasets. Unlike other text-generation tasks, we observe that LLM-based judges perform quite poorly (usually below 50\% accuracy) in the biomedical relation extraction task. Our findings reveal that it happens mainly because relations extracted by LLMs do not adhere to any standard format. 
To address this, we propose structured output formatting for LLM-generated responses that helps \textit{LLM-Judges} to improve their performance by about 15\% (on average). We also introduce a domain adaptation technique to further enhance  \textit{LLM-Judge} performance by effectively transferring knowledge between datasets. 
We release both our human-annotated and LLM-annotated judgment data (36k samples in total) here: \url{https://github.com/tahmedge/llm_judge_biomedical_re}.
\end{abstract}

\section{Introduction}
\begin{figure}
    \centering
    \includegraphics[width=\linewidth]{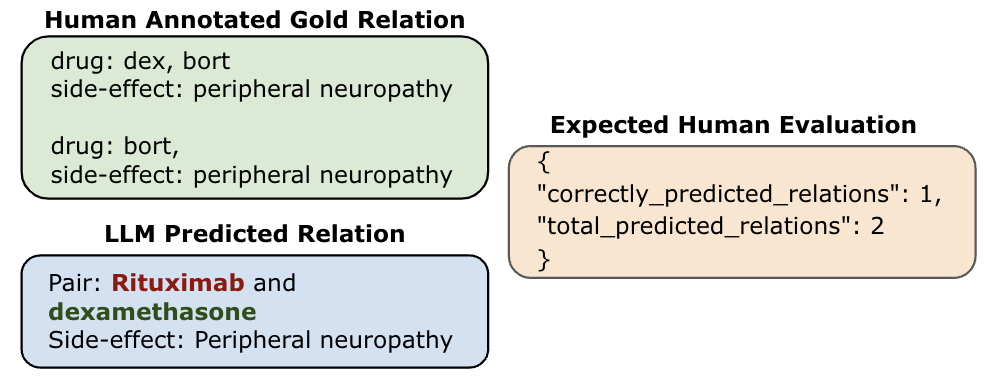}
    \caption{{An example of human evaluation of LLM-generated outputs in comparison to the gold labels. The drug ``dex'' in the gold relation is predicted as ``dexamethasone'' in LLM response. While exact match will fail in this case, human evaluators can recognize.}} 
    \label{fig:human_eval_example}
\end{figure}
{Relation extraction, the task of identifying meaningful associations between biomedical entities such as drugs, diseases, and genes from vast amounts of unstructured text \cite{bassignana-plank-2022-mean}, is a cornerstone of biomedicine. As the volume of unstructured biomedical text grows exponentially, efficient and accurate relation extraction is no longer a convenience but a necessity for advancing medical research, drug discovery, and improving patient outcomes \cite{luo2022biogpt}. Recent research has shown that Large Language Models (LLMs) can achieve strong performance in biomedical relation extraction, even in zero-shot scenarios \cite{jahan2024comprehensive}. This capability makes LLMs particularly valuable for real-world biomedical applications where manually annotated datasets are scarce or costly to obtain.}

However, 
{LLM's ability to generate human-like text introduces a major challenge in evaluation \cite{laskar-etal-2024-systematic,jahan2024comprehensive}. Traditional automatic evaluation methods, such as string matching and token overlap, fail to capture semantic equivalence, as LLMs frequently produce synonyms, abbreviations, or paraphrased responses that are meaningfully correct but not exact matches. For example, an LLM may generate ``Hepatic carcinoma'' instead of ``Liver cancer'', leading conventional metrics to misclassify correct extractions as incorrect. Due to these limitations, human evaluation has been the predominant method (see Figure \ref{fig:human_eval_example} for an example) for assessing LLM performance in biomedical relation extraction \cite{jahan2024comprehensive}. However, human evaluation is slow and costly, making it impractical for real-world applications.}


{To address this, LLMs-as-the-Judge \cite{zheng2023judging} have gained attention as a potential alternative to human evaluation. While this paradigm has been explored in general NLP tasks \cite{li2024llms,gu2024surveyllmasajudge}, there is currently no biomedical-specific \textit{LLM-Judge} designed to evaluate relation extraction tasks. Unlike open-domain text generation, biomedical relation extraction requires precise domain knowledge, standardized terminology, and strict adherence to extract relationships between entities. This complexity raises concerns about whether existing \textit{LLM-Judges} can reliably assess biomedical extractions.}

{To investigate this, we examined the capability of several LLMs-as-Judges in evaluating the responses generated by different LLMs (\textit{LLM-Generators}) across multiple biomedical relation extraction datasets \cite{jahan2024comprehensive}. Surprisingly, despite prior success in general NLP evaluation tasks, \textit{LLM-Judges} performed very poorly in biomedical relation extraction in comparison to human evaluators (Figure \ref{fig:average_cibm}). These findings suggest that domain specificity may significantly impact the effectiveness of LLMs-as-Judge, underscoring the need for adaptation in biomedicine. 

\begin{figure}
    \centering
    \includegraphics[width=0.9\linewidth]{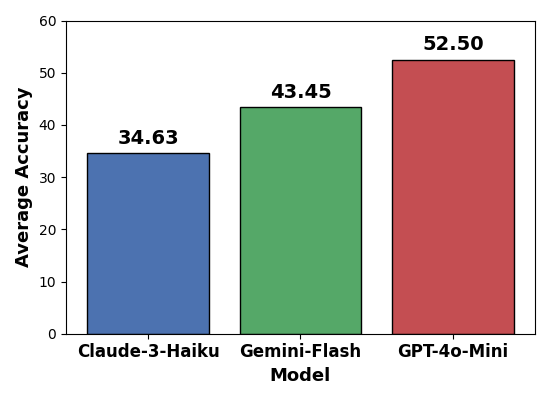}
    \caption{Average judgment accuracy based on our evaluation of different LLMs-as-the-judge across 3 Relation Extraction datasets (KD-DTI, DDI, BC5CDR) to evaluate the LLM-generated responses in \citet{jahan2024comprehensive}.} 
    \label{fig:average_cibm}
\end{figure}

To address the shortcomings of {LLMs-as-Judge} in the biomedical relation task, we propose \textit{structured output format} \cite{xia2024fofo,li-etal-2024-simple} for response generation by \textit{LLM-Generators}. Based on extensive experiments, we find that structured output format in the response generated by \textit{LLM-Generators} consistently helps the \textit{LLM-Judges} to improve their performance 
across various relation extraction datasets. 

Moreover, we also find that there is a lack of human-annotated judgment data that prohibits the training of \textit{LLM-Judges} for relation extraction. Therefore, we propose a domain adaptation \cite{laskar2022domain} technique to address the lack of human-annotated judgment data by effectively transferring knowledge from out-of-domain data to improve the performance of \textit{LLM-Judges} in the target domain. 
Our major contributions are summarized below:

\begin{itemize}
    \item We provide the first comprehensive study of \textit{LLM-Judges} in biomedical relation extraction, benchmarking 8 \textit{LLM-Judges} on responses generated by 5 \textit{LLM-Generators} across 3 biomedical relation extraction datasets. Our findings demonstrate that LLMs are not reliable to serve as evaluators in biomedicine, highlighting their significant performance gap compared to human evaluators.
    \item To address the above limitation, we propose structured output formatting in LLM-generated responses for biomedical relation extraction to improve the performance of \textit{LLM-Judge}. We also propose a domain adaptation technique to effectively transfer knowledge from one domain to another to further improve \textit{LLM-Judge} performance in biomedical relation extraction. 
    \item We conduct over 100 
    experiments, analyzing the impact of structured output format, domain adaptation, and model scaling on \textit{LLM-Judge} performance. These experiments reveal critical insights into 
    why LLM-based evaluation fails in biomedical relation extraction, establishing the need for task-specific evaluation frameworks. 
    \item We make our judgment data (4k human-annotated and 32k LLM-annotated samples) consisting of 3 relation extraction datasets publicly available\footnote{\url{https://github.com/tahmedge/llm_judge_biomedical_re}}.
\end{itemize}


\begin{figure*}[t!]
    \centering
    \includegraphics[width=\linewidth]{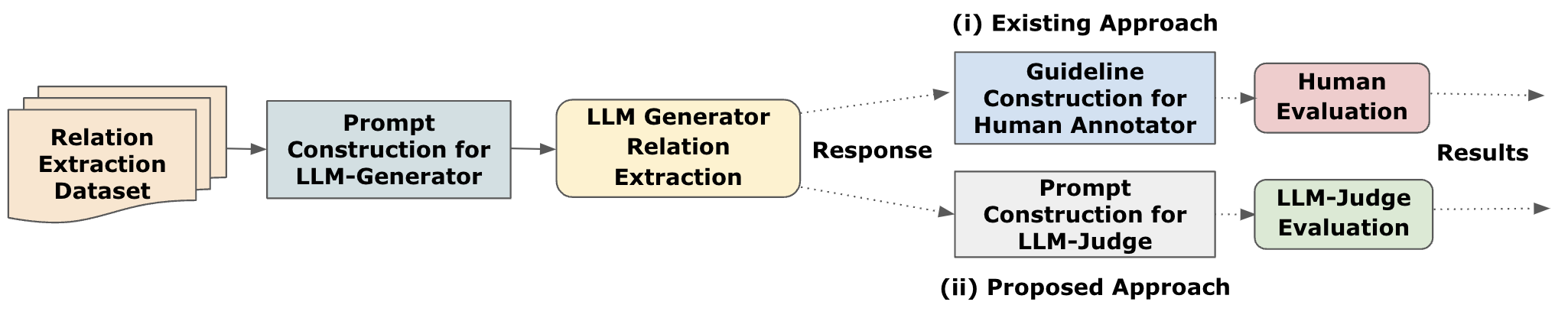}
    \caption{{An Overview of LLM-based  Relation Extraction System. After constructing the input prompt for the \textit{LLM-Generator} for a given dataset, the \textit{LLM-Generator} generates a response. The evaluation requires the identification of \textit{\textbf{the number of correctly predicted relations}} and  \textit{\textbf{the total number of predicted relations}} in the response to evaluate the performance of the \textit{LLM-Generator} for relation extraction in terms of metrics like Precision, Recall, and F1. 
    }}
    \label{fig:overview}
\end{figure*}

\section{Background}

\subsection{Relation Extraction Task Description}

The relation extraction task aims to extract relations between
named entities in a given text \cite{zhong2021frustratingly-re}. The biomedical relation extraction task aims to identify relationships between biomedical named entities like genes, drugs, and diseases \cite{chen2023extensive}. More specifically, in the context of biomedicine, the goal is to analyze textual data to identify which gene/variant is responsible or which treatment/drug is effective for which disease, as well as identifying drug-drug interactions, and etc. An example for disease-treatment relation extraction is given below. 
  
\noindent  \textit{\textbf{Example Text:} The patient has been given chemotherapy for their rare form of cancer.}

\noindent  \textit{\textbf{Expected Relation:} ``Chemotherapy'' is a treatment for ``rare form of cancer''.}

\subsection{Related Work}

Traditional approaches for biomedical relation extraction relied on supervised learning techniques that required large, manually annotated datasets \cite{luo2022biogpt}. However, the construction of such datasets is expensive and time-consuming. Recently, LLMs have demonstrated impressive zero-shot performance across a wide range of NLP tasks \cite{laskar-etal-2023-systematic,bang2023multitaskchatgpt,qin2023chatgpt}. This has inspired researchers to successfully apply LLMs for biomedical relation extraction, particularly in zero-shot scenarios where annotated data is scarce \cite{jahan-etal-2023-evaluation,jahan2024comprehensive,tian2024opportunities}. Despite LLMs achieving impressive zero-shot performance in the biomedical relation extraction task, evaluating the performance of LLM in this task remains challenging \cite{jahan2024comprehensive}. This is because LLMs may generate valid outputs that are semantically equivalent to the gold standard but differ in surface form (e.g., synonyms, abbreviations). This issue is further exacerbated in biomedicine due to the nuanced and complex nature of biomedical language.  Therefore, existing automatic metrics like string matching or n-gram overlap often fall short in assessing the semantic correctness of LLM-generated free-form responses \cite{laskar-etal-2024-systematic}.



In the context of biomedical relation extraction, while human experts can assess the relevance of LLM extracted relations \cite{jahan2024comprehensive}, human evaluation is inherently time-consuming, expensive, and lacks scalability. This makes it impractical for rapid iteration cycles in research and real-world deployment scenarios. The paradigm of using LLMs-as-judges \cite{zheng2023judging} for evaluating free-form text generated by other LLMs has recently gained a lot of attention as a potential alternative to human evaluations \cite{laskar-etal-2024-systematic}. While prior research suggests that LLM-based evaluators can capture linguistic nuances often overlooked by traditional metrics \cite{li2024llms,gu2024surveyllmasajudge}, most of these studies have focused on general-domain tasks, without exploring their effectiveness in specialized domains such as biomedical relation extraction. To this end, this paper explores the potential of LLMs-as-the-Judge for evaluating the response generated by other LLMs in the biomedical relation extraction task to mitigate the dependence on expensive, time-consuming human evaluation, alongside ensuring reliability in LLM evaluation by capturing domain-specific nuances that traditional metrics may overlook. 


\section{Methodology} 
\subsection{LLM-based Relation Extraction Systems} 
We show the overall pipeline to develop an LLM-based end-to-end relation extraction system 
in Figure \ref{fig:overview}. Here, at first, for a given relation extraction dataset, pre-processing steps are first applied to construct a prompt. This prompt is then provided to an LLM (i.e., \textit{LLM-Generator}), which extracts relations. Afterwards, we demonstrate two evaluation paradigms which we describe below:

(i) \textit{\textbf{Existing Approaches based on Human Evaluation:}} Where an annotation guideline is first constructed to evaluate the performance of the \textit{LLM-Generator} (i.e., relation extraction LLM) by human annotators. The  LLM-generated response is then sent to the human annotators for evaluation. 

(ii) \textit{\textbf{Our Proposed Approach based on LLM-Judge:}} Where the prompt for the \textit{LLM-Judge} is first constructed to evaluate the performance of the \textit{LLM-Generator}. The response generated by the \textit{LLM-Generator} is then evaluated by the \textit{LLM-Judge}. This is our proposed approach where human evaluator(s) are replaced with the \textit{LLM-Judge}. 
In the following, we describe our proposed \textit{LLM-Judge} to evaluate relation extraction models.

\subsection{LLM-Judge for LLM-based Relation Extraction Model Evaluation}

\paragraph{Prompting:} 
To utilize \textit{LLM-Judge} for the evaluation of LLM-generated responses (i.e., \textit{LLM-Generators}) in relation extraction, we first design the prompt for the judge, as described below. 
\definecolor{attachedColor}{HTML}{e0efff}
\definecolor{attachedColor2}{HTML}{f3f3f3}
\begin{tcolorbox}[
boxrule=0.25pt,   
  colback=attachedColor2,    
  colframe=black,           
  colbacktitle=attachedColor, 
  coltitle=black,           
  title={Prompt to the LLM-Judge},
  fonttitle=\bfseries,      
  fontupper=\small          
]
{You are required to annotate the response generated by an AI model for biomedical relation extraction.} You are given the \textbf{relation extraction task description}, followed by the \textbf{biomedical text}, then the \textbf{human annotated gold relations}, and finally the \textbf{AI model predicted relations}. Now, identify how many of the AI predicted relations are correct in comparison to the gold relations. Also, identify how many relations in total are predicted by AI. Generate your response in the JSON format with the following keys:
\begin{enumerate}
    \item \texttt{correctly\_predicted\_relations}
    \item \texttt{total\_predicted\_relations}
\end{enumerate}

\textbf{[TASK DESCRIPTION]} \\
\textbf{[BIOMEDICAL TEXT]} \\
\textbf{[GOLD RELATIONS]} \\
\textbf{[AI PREDICTED RELATIONS]}
\end{tcolorbox}

We designed this prompt based on extensive prompt engineering 
using several LLMs (e.g., ChatGPT, Gemini, Claude, etc.) on the outputs generated by different \textit{LLM-Generators} in \citet{jahan2024comprehensive} across various relation extraction datasets. 

\paragraph{Structured Output Formatting:} During prompt engineering in our previous step, we notice that \textit{LLM-Judges} often find it difficult to properly understand the response generated by various \textit{LLM-Generators} since the generated responses are mostly unstructured text. To address this issue, we propose \textit{Structured Output Formatting} for the \textit{LLM-Generators}. In our proposed \textit{Structured Output Formatting} approach, we require the \textit{LLM-Generators} to generate their response in a structured format, i.e., JSON format (see Figure \ref{fig:u_vs_s} for an example).  JSON format was selected since recent research has demonstrated that LLMs are more reliable in generating responses in ``JSON'' instead of other formats like ``YAML'' \cite{laskar-etal-2024-query}.

\begin{figure}
    \centering
    \includegraphics[width=\linewidth]{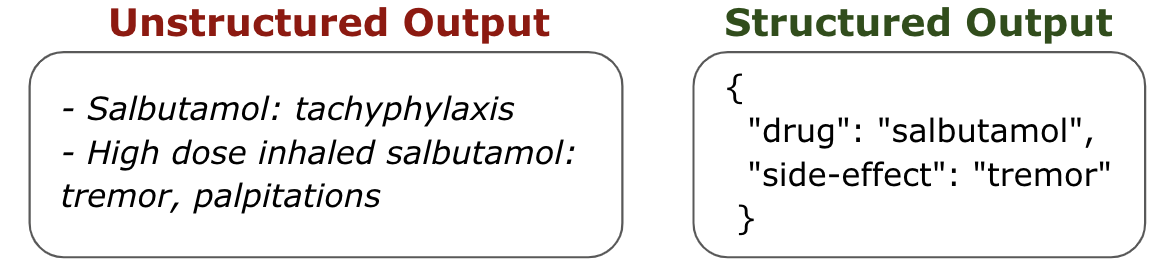}
    \caption{{An example of the LLM-generated outputs for \textit{drug} and \textit{drug-induced side-effects} relation extraction in the \textit{unstructured} format used by \citet{jahan2024comprehensive} and our proposed \textit{structured} format.}}
    \label{fig:u_vs_s}
\end{figure}

\paragraph{Domain Adaptation via Transfer Learning:} Leveraging closed-source LLMs-as-the-judge have several limitations. For instance, closed-source LLMs are continuously updated. Therefore, the release of an updated version of a closed-source model often makes their respective earlier versions obsolete \cite{DBLP:journals/corr/abs-2405-14782,laskar-etal-2024-systematic}. This creates a huge reproducibility concern. While open-source LLMs could be a suitable alternative, they do not have the superior zero-shot ability of their closed-source counterparts \cite{jahan2024comprehensive}. Moreover, we find that existing open-source \textit{LLM-Judges} like Prometheus-2 \cite{kim2024prometheus2} are restricted to certain evaluation dimensions (e.g., pairwise ranking or pointwise scoring for specific criteria: helpfulness, factual correctness, etc.) since they are fine-tuned only for such qualitative metrics. This prohibits the customization of the judging criteria based on user needs, 
making them inapplicable in our case to evaluate the relation extraction performance of other LLMs. 

Furthermore, due to the absence of human evaluation data in existing relation extraction datasets for LLM-generated responses, it is difficult to train a reliable \textit{LLM-Judge} model to 
evaluate the response generated by different relation extraction models. To address this low-resource problem that prohibits the training of \textit{LLM-Judge} for relation extraction, we propose a domain adaptation technique  \cite{garg2020tanda,laskar2022domain}. In our proposed approach, the \textit{LLM-Judge} is fine-tuned on a limited number of human annotated out-of-domain relation extraction data to make it more specialized for relation extraction evaluation. More specifically, suppose we have two datasets, $X$ and $Y$, where the dataset $X$ may focus on \textit{drug-drug-interaction} extraction and the dataset $Y$ may focus on \textit{disease-treatment-relation} extraction. In our domain adaptation technique, if the target dataset for the judge model to evaluate is the dataset $X$, then we first fine-tune the judge model on the dataset $Y$ if we have some human evaluation data (e.g., human-annotated judgment labels containing the number of predicted relations and total predicted relations by the \textit{LLM-Generator}) available for the dataset $Y$. In this way, we specialize the \textit{LLM-Judge} for biomedical relation extraction evaluation.

\section{Experiments}
To evaluate our proposed solution of replacing human evaluators with \textit{LLM-Judges}, we conduct extensive experiments with various {LLMs-as-the-Judge} on the responses generated by different \textit{LLM-Generators} for biomedical relation extraction. 
Below, we first describe the datasets used in our experiments. Then, we describe the LLMs used as the \textit{Judge} and the \textit{Generator}. Afterwards, we present our evaluation metrics to measure the performance of the \textit{LLM-Judge}. Finally, we briefly describe the implementation details. 

\subsection{Datasets}

To evaluate the \textit{LLM-Judges}, we use the responses generated by the \textit{LLM-Generators} for \textit{chemical-disease-relation} extraction ($500$ test samples) in the {BC5CDR} dataset \cite{bc5cdr}, \textit{drug-target-interaction} extraction ($1159$ test samples) in the {KD-DTI} dataset \cite{kddti}, and \textit{drug-drug-interaction} extraction ($191$ test samples) in the {DDI} dataset \cite{ddi}.

\subsection{Models}
\paragraph{LLM-Generators:}
To evaluate the performance of different \textit{LLM-Judges}, at first we benchmark their performance on different LLM-generated responses
collected from \citet{jahan2024comprehensive}. 
The responses from the following \textit{LLM-Generators} are used: (i) GPT-3.5 \cite{openai2023gpt4}, (ii) Claude-2 \cite{anthropicclaude2}, (iii) PaLM-2 \cite{anil2023palm2}, (iv) LLaMA-2-13B \cite{touvron2023llama2}. Moreover, to investigate the effectiveness of \textit{Structured Output Format}, we regenerate the response using {GPT-4-Turbo} \cite{openai2023gpt4} in both unstructured and structured format, and compare the performance of the \textit{LLM-Judge} when the response is generated in different formats by the same \textit{LLM-Generator}. For this case, we did not use the \textit{LLM-Generators} used in \citet{jahan-etal-2023-evaluation} since they are early generation LLMs and we observe that they fail to generate the response in structured format.

\paragraph{LLM-Judges:}

For the \textit{LLM-Judges}, we select the cheapest versions of the currently available closed-source LLMs (GPT-4, Gemini, and Claude-3) to minimize their usage cost. For the open-source \textit{LLM-Judges} (LLaMA, Phi, Qwen, and DeepSeek), we select the models with less than 10B parameters since they can be used in a machine with just 1 L4 GPU \cite{fu2024tiny}, making them cost-effective for real-world deployment. More specifically, for the \textit{LLM-Judge}, we primarily use the following LLMs: 
(i) GPT-4o-Mini \cite{openai2023gpt4}, (ii) Gemini-1.5-Flash \cite{team2023gemini}, (iii) Claude-3-Haiku \cite{anthropicclaude3}, (iv) LLaMA-3.1-8B-Instruct\footnote{\url{https://huggingface.co/meta-llama/Llama-3.1-8B-Instruct}} \cite{dubey2024llama3}, (v) Qwen-2.5-7B-Instruct\footnote{\url{https://huggingface.co/Qwen/Qwen2.5-7B-Instruct}} \cite{yang2024qwen2.5}, and (vi) Phi-3.5-Mini-3.8B-Instruct\footnote{\url{https://huggingface.co/microsoft/Phi-3.5-mini-instruct}} \cite{abdin2024phi}. 
With the recent success of reasoning-based LLMs like DeepSeek-R1 \cite{guo2025deepseek}, we also use its distilled versions based on Qwen and LLaMA, (vii) DeepSeek-R1-Distill-Qwen-7B\footnote{\url{https://huggingface.co/deepseek-ai/DeepSeek-R1-Distill-Qwen-7B}} and (viii) DeepSeek-R1-Distill-Llama-8B\footnote{\url{https://huggingface.co/deepseek-ai/DeepSeek-R1-Distill-Llama-8B}}, respectively. Appendix \ref{appnedix_models} contains detailed model descriptions.

While we also tried the biomedical domain-focused BioMistral-7B model \cite{labrak2024biomistral} as a judge, we observed that it failed to follow the judging instruction and could not evaluate any \textit{LLM-Generators} response (see Appendix
\ref{biomistral-output} for details). Therefore, we did not report its results.

\subsection{Evaluation Metrics}

We use the following metrics to evaluate the effectiveness of the \textit{LLM-Judge}.

\paragraph{Exact Match Accuracy:} Measures the Exact Match Accuracy of the \textit{LLM-Judge} annotation in comparison to the human-annotated 
gold label. For \textit{exact match} calculation, if there is a match for both the number of \textit{correctly predicted relations} and \textit{total predicted relations} between the gold human annotation and the \textit{LLM-Judge} annotation, then we consider it as an exact match, and the score will be 1. Otherwise, the score will be 0. Thus, a higher exact match score denotes better performance. 

\paragraph{Root Mean Squared Error:} Measures the Root Mean Squared Error (RMSE) distance between the \textit{LLM-Judge} annotation and the human-annotated gold label. Mean Squared Error (MSE) is defined as the average of the squared differences between the predictions (in this case, the \textit{LLM-Judge} annotations) and the actual values (the \textit{Human-Judge} annotated gold labels). RMSE, which is the square root of the MSE, helps penalize larger errors more severely while being in the same units as the target variable. 
Suppose the \textit{correctly predicted relations} and the \textit{total predicted relations} annotated by humans are \textcolor{blue}{0} and \textcolor{blue}{2}, respectively, and by \textit{LLM-Judge} are \textcolor{purple}{1} and \textcolor{purple}{2}, respectively, then the RMSE distance:

\[
\sqrt{\frac{1}{2} \left( (\textcolor{purple}{1} - \textcolor{blue}{0})^2 + (\textcolor{purple}{2} - \textcolor{blue}{2})^2 \right)}
= \sqrt{0.5}
\approx 0.71.
\]
Contrary to the exact match score, a lower RMSE distance denotes better performance. 

\begin{table*}[t!]
\centering
\scriptsize
\resizebox{0.85\textwidth}{!}{
\centering
\begin{tabular}{lcc|cc|cc}
\toprule
& \multicolumn{2}{c|}{\textbf{BC5CDR}} & \multicolumn{2}{c|}{\textbf{DDI}} & \multicolumn{2}{c}{\textbf{KD-DTI}} \\
\cmidrule(lr){2-7}
           \textbf{LLM-Judge}         & \textbf{EM ($\Delta$)} & \textbf{RMSE ($\nabla$)} & \textbf{EM ($\Delta$)} & \textbf{RMSE ($\nabla$)} & \textbf{EM ($\Delta$)} & \textbf{RMSE ($\nabla$)} \\
\midrule
\multicolumn{1}{l|}{Phi-3.5-Mini-3.8B-Instruct}      & 33.80 & 2.40  & 43.06 & 2.19   & 35.55 & 2.11 \\
\multicolumn{1}{l|}{Qwen-2.5-7B-Instruct}             & 45.25 & 2.42  & 46.60 & 2.15   & 49.98 & 1.82 \\
\multicolumn{1}{l|}{LLaMA-3.1-8B-Instruct}             & 29.45 & 2.40  & 29.32 & 2.95   & 36.73 & 2.10 \\
\midrule
\multicolumn{1}{l|}{Deepseek-R1-Qwen-7B}              & 30.60 & 2.76  & 42.67 & 3.07   & 42.45 & 2.51 \\
\multicolumn{1}{l|}{Deepseek-R1-LLaMA-8B}             & 30.50 & 3.37  & 42.15 & 4.16   & 33.48 & 3.25 \\
\midrule
\multicolumn{1}{l|}{Claude-3-Haiku}                  & 29.50 & 2.26  & 31.15 & 2.70   & 40.27 & 1.83 \\
\multicolumn{1}{l|}{Gemini-Flash}                    & 42.55 & \textbf{2.09}  & 47.12 & 2.11   & 40.68 & 1.98 \\
\multicolumn{1}{l|}{GPT-4o-Mini}                     & \textbf{48.35} & 2.33  & \textbf{59.03} & \textbf{1.84}   & \textbf{53.11} & \textbf{1.81} \\
\bottomrule
\end{tabular}}
\caption{Performance of different \textit{LLM-Judges} on the responses generated by the \textit{LLM-Generators} in \citet{jahan2024comprehensive} across three datasets: \textbf{BC5CDR}, \textbf{DDI}, and \textbf{KD-DTI}. The \textbf{Exact Match (EM)} Accuracy (higher is better, indicated by $\Delta$) and the \textbf{Root Mean Squared Error (RMSE)} (lower is better, indicated by $\nabla$) are reported. The reported score for each \textit{LLM-Judge} is the average of their evaluations for all \textit{LLM-Generators} within each dataset.} 
\label{tab:cibm_dataset_performance_all_models}
\end{table*}

\begin{table*}[t]
\small
\setlength{\tabcolsep}{2pt} 
\resizebox{\textwidth}{!}{
\centering
\begin{tabular}{lcc|cc|cc|cc|cc|cc}
\toprule
 & \multicolumn{4}{c}{\textbf{BC5CDR}} & \multicolumn{4}{c}{\textbf{DDI}} & \multicolumn{4}{c}{\textbf{KD-DTI}} \\
\cmidrule(lr){2-5} \cmidrule(lr){6-9} \cmidrule(lr){10-13}
 & \multicolumn{2}{c}{\textbf{Structured}} & \multicolumn{2}{c}{\textbf{Unstructured}} 
 & \multicolumn{2}{c}{\textbf{Structured}} & \multicolumn{2}{c}{\textbf{Unstructured}} 
 & \multicolumn{2}{c}{\textbf{Structured}} & \multicolumn{2}{c}{\textbf{Unstructured}} \\
\cmidrule(lr){2-3} \cmidrule(lr){4-5} \cmidrule(lr){6-7} \cmidrule(lr){8-9} \cmidrule(lr){10-11} \cmidrule(lr){12-13}
\textbf{LLM-Judge} 
  & \textbf{EM ($\Delta$)} & \textbf{RMSE ($\nabla$)}  
  & \textbf{EM ($\Delta$)} & \textbf{RMSE ($\nabla$)}  
  & \textbf{EM ($\Delta$)} & \textbf{RMSE ($\nabla$)}  
  & \textbf{EM ($\Delta$)} & \textbf{RMSE ($\nabla$)}  
  & \textbf{EM ($\Delta$)} & \textbf{RMSE ($\nabla$)}  
  & \textbf{EM ($\Delta$)} & \textbf{RMSE ($\nabla$)}  \\
\midrule
\multicolumn{1}{l|}{Phi-3.5-Mini-3.8B-Instruct}
  & 53.80 & 0.94 & 35.60 & 2.74 
  & 32.89 & 1.92 & 32.51 & 2.42 
  & 36.94 & 2.02 & 36.74 & 2.55 \\
  
\multicolumn{1}{l|}{Qwen-2.5-7B-Instruct}
  & 67.80 & 0.93 & 44.60 & 2.84 
  & 43.93 & 1.64 & 43.46 & 2.08 
  & 61.78 & 1.19 & 55.82 & 2.18 \\
  
\multicolumn{1}{l|}{LLaMA-3.1-8B-Instruct}
  & 59.20 & 0.85 & 33.00 & 2.67 
  & 37.70 & 1.99 & 30.37 & 2.43 
  & 40.38 & 2.44 & 38.22 & 2.46 \\
\midrule
\multicolumn{1}{l|}{Deepseek-R1-Qwen-7B}
  & 57.00 & 1.28 & 31.60 & 3.32 
  & 41.88 & 2.55 & 35.60 & 2.85 
  & 47.71 & 1.83 & 46.16 & 2.65 \\
  
\multicolumn{1}{l|}{Deepseek-R1-LLaMA-8B}
  & 54.40 & 2.25 & 37.40 & 3.64 
  & 37.70 & 4.14 & 35.60 & 4.62 
  & 48.40 & 2.68 & 42.36 & 3.31 \\
\midrule
\multicolumn{1}{l|}{Claude-3-Haiku}
  & 67.80 & \textbf{0.60} & 36.80 & 2.67 
  & 32.46 & 3.01 & 25.13 & 2.80 
  & 59.28 & 1.22 & 50.04 & 2.22 \\
\multicolumn{1}{l|}{Gemini-1.5-flash} 
  & \textbf{70.20} & 0.66 & 48.40 & 2.75 
  & 43.98 & 1.75 & 39.27 & 2.29 
  & 41.59 & 1.68 & 41.52 & 2.47 \\
\multicolumn{1}{l|}{GPT-4o-mini}
  & 67.20 & 0.73 & 43.20 & 2.72 
  & \textbf{53.93} & \textbf{1.45} & 47.12 & 2.14 
  & \textbf{72.39} & \textbf{1.08} & 66.35 & 2.13 \\
\bottomrule
\end{tabular}}
\caption{Performance Comparison between \textit{\textbf{Structured}} (our proposed) and \textit{\textbf{Unstructured}} (baseline) output format. All the responses are generated using GPT-4-Turbo as the \textit{LLM-Generator}.}
\label{tab:s_vs_u_results}
\end{table*}

\begin{table}[t]
\scriptsize
\setlength{\tabcolsep}{4pt} 
\centering
\begin{tabular}{lc|c|l}
\toprule

& \multicolumn{2}{c}{\textbf{Data}} &  \\ \cmidrule(lr){2-3}
\textbf{LLM-Judge} & \textbf{Fine-Tuning} & \textbf{Evaluation} & \textbf{EM Accuracy} \\

\midrule
\multicolumn{1}{l|}{Qwen-2.5-7B-Instruct} & {BC5CDR} & {KD-DTI} & 75.75 (+13.97) \\
\multicolumn{1}{l|}{Qwen-2.5-7B-Instruct} & {KD-DTI} & {BC5CDR} & 71.40 (+3.60)   \\
\midrule 
\multicolumn{1}{l|}{Phi-3.5-Mini-3.8B-Instruct} & {BC5CDR} & {KD-DTI} & 69.54 (+32.60) \\
\multicolumn{1}{l|}{Phi-3.5-Mini-3.8B-Instruct} & {KD-DTI} & {BC5CDR}   & 64.80 (+11.00)  \\

\bottomrule
\end{tabular}
\caption{Effectiveness of Domain Adaptation via Transfer Learning. The value in brackets indicates the performance gain relative to the Zero-Shot results for \textit{Structured} output in respective evaluation datasets in Table \ref{tab:s_vs_u_results}.}
\label{tab:ft_only}
\end{table}

\subsection{Implementation}
For the inference of \textit{LLM-Generators} and \textit{LLM-Judges}, we use the temperature value of 1.0, with other decoding parameters being set to the default values: as given in the respective API providers (OpenAI\footnote{\url{https://platform.openai.com/docs/api-reference/introduction}}, Google\footnote{\url{https://ai.google.dev/gemini-api/docs}}, Anthropic\footnote{\url{https://www.anthropic.com/api}})  for the closed-source models, and in  HuggingFace
\cite{wolf2020transformers} for the open-source models. We select the temperature value of 1.0 to allow more diversity in LLM-generated responses  such that it allows us to ensure a more robust evaluation of the \textit{LLM-Judges} in diverse output scenarios.  Since both the \textit{LLM-Generators} and the \textit{LLM-Judges} do not need to produce outputs of longer sequence length, the maximum output tokens was set to 300 tokens except for the reasoning models (DeepSeek-R1). For the DeepSeek-R1-based models, we increased the output token limit to 1000 tokens to allow the model enough tokens for thinking. For domain adaptation via transfer learning, we fine-tune for $3$ \textit{epochs} with the \textit{batch size} = $1$, \textit{learning rate} = $2e-5$, and \textit{sequence length} = $3k$. All experiments were run using 1 A100 GPU. 

\section{Results and Discussion}

\subsection{Performance in Existing LLM-based Relation Extraction Benchmarks}

In this section, we benchmark the performance of different \textit{LLM-Judges} to evaluate the responses generated by different \textit{LLM-Generators} used by \citet{jahan2024comprehensive}. In their work, \citet{jahan2024comprehensive} used GPT-3.5 \cite{openai2023gpt4}, PaLM-2 \cite{anil2023palm2}, Claude-2 \cite{anthropicclaude2}, and LLaMA-2-13B-Instruct \cite{touvron2023llama2} as the \textit{LLM-Generators}. We collected their human evaluator annotated labels consisting of the number of \textit{correctly predicted relations} and \textit{total predicted relations} for each \textit{LLM-Generator}. Then we measure the Exact Match Accuracy and the RMSE Distance between the \textit{LLM-Judge} annotation (the number of \textit{correctly predicted relations} and the \textit{total predicted relations} annotated by the \textit{LLM-Judge}) and the human annotation.

For each dataset, we then compute the average Exact Match Accuracy and RMSE Distance across all \textit{LLM-Generators}, as demonstrated in Table \ref{tab:cibm_dataset_performance_all_models}. From  Table \ref{tab:cibm_dataset_performance_all_models}, we find that none of the \textit{LLM-Judges} could reach accuracy above 50\%, except GPT-4o-Mini. Nonetheless, GPT-4o-Mini still fails to achieve more than 60\% accuracy in any datasets. Our hypothesis is that the responses generated by the \textit{LLM-Generators} used by 
 \citet{jahan2024comprehensive} were quite unstructured, which could be difficult for the \textit{LLM-Judges} to evaluate. In the following section, we investigate whether \textit{Structured Output Formatting} could alleviate this issue.

 \subsection{Performance of LLM-Judges based on Structured vs Unstructured Output}\label{sec:structured}
 To improve the \textit{LLM-Judge} accuracy, we generate the responses in all three datasets using the {GPT-4-Turbo} \cite{openai2023gpt4} model as the generator, with specifically prompting {GPT-4-Turbo} to extract the relations between the named entities in a more structured way, i.e., JSON format. To compare whether this structured output format could improve the performance of the \textit{LLM-Judges}, we also regenerate the responses without any structured output format instruction by using the same prompt as used by \citet{jahan2024comprehensive}. Afterward, two human annotators having backgrounds in NLP and biomedicine annotated the \textit{correctly predicted relations} and the \textit{total predicted relations} in the GPT-4-Turbo generated responses for both structured and unstructured cases. When discrepancies arise between annotations from different annotators, they are resolved through discussions. 
 
 We then evaluate the performance of \textit{LLM-Judges} in these structured responses as well as the unstructured responses generated by GPT-4-Turbo. Based on the results presented in Table \ref{tab:s_vs_u_results}, we find that structured formatting consistently improves the performance for all models. While in the BC5CDR dataset, {Gemini-1.5-Flash} achieves the best performance in terms of accuracy with {Claude-3-Haiku} achieving the best result in terms of the RMSE metric, in other two datasets (DDI and KD-DTI), GPT-4o-Mini achieves the best performance in terms of both metrics. For open-source LLMs, in the BC5CDR dataset, {Qwen-2.5-7B} has the best accuracy and {LLaMA-3.1-8B} performs the best in terms of RMSE distance. However, the accuracy is quite low for {LLaMA-3.1-8B} while being larger than {Qwen-2.5-7B}. In the other two datasets (DDI and KD-DTI), Qwen-2.5-7B-Instruct achieves the best result in terms of both metrics among open-source models, even outperforming reasoning-based DeepSeek-Distilled models. 

For all three datasets, we find based on paired \(t\)-test that the performance difference between the \textit{structured} and \textit{unstructured} approach is \textbf{statistically significant} \((p < 0.05)\) for both metrics.

 
\subsection{Effectiveness of Domain Adaptation via Transfer Learning}

In our prior experiments, we demonstrate that proprietary LLMs demonstrate better performance as the judge in zero-shot scenarios. However, issues such as transparency, reproducibility, and cost highlight the need for open-source \textit{LLM-Judge}. 
While early work \cite{kim2023prometheus,kim2024prometheus2} demonstrates the effectiveness of training an open-source \textit{LLM Judge}, to our best knowledge, there is no open-source \textit{LLM-Judge} trained on biomedical data currently available. Since there is also a lack of
human-annotated judgment data that prohibits the
training of \textit{LLM-Judges} for relation extraction, we investigate a domain adaptation technique via transfer learning to address this low-resource problem.

Recent research has demonstrated that language models can effectively transfer knowledge from one dataset to another \cite{garg2020tanda,laskar2022domain}. Inspired by the idea, in this work, we also propose transfer learning from one relation extraction dataset to the other. Based on our human-annotated labels for the \textit{Structured} output in Section \ref{sec:structured}, we investigate two scenarios: (i) Fine-Tuning on the BC5CDR dataset (500 samples) and Evaluation on the KD-DTI dataset (1159 samples), and (ii) Fine-Tuning on the KD-DTI dataset (1159 samples) and Evaluation on the BC5CDR dataset (500 samples). For hyperparameter tuning, we use the DDI dataset as the validation set. 
We fine-tune the Qwen-2.5-7B-Instruct and Phi-3.5-3.8B-Instruct models because our previous experiments demonstrated that Qwen with 7 billion parameters outperforms the larger 8-billion parameter LLaMA model, while Phi-3.5-3.8B achieves comparable performance with the 2x larger LLaMA model. 

Based on the results presented in Table \ref{tab:ft_only}, we find that our proposed domain adaptation strategy by transferring knowledge via fine-tuning on a limited amount of labeled data could significantly improve the accuracy. The performance gains in BC5CDR and KD-DTI by Qwen-2.5-7B-Instruct even outperform the closed-source models in Table \ref{tab:s_vs_u_results}. This demonstrates the effectiveness of our domain adaptation technique in low-resource scenarios. 

\subsection{Impact of Model Scaling}

Since one of our motivations behind \textit{LLM-Judge} is to reduce cost and improve efficiency to ensure their real-world utilization, we primarily selected open-source models having less than 10B parameters or the most cost-efficient version of different closed-source LLM providers. In this section, we investigate two scenarios: 

(i) What is the impact of reducing the size of the best-performing open-source model? 

(ii) Can domain adaptation via fine-tuning help even smaller models outperform larger models?

Below, we demonstrate our findings.

\begin{figure}[t!]
    \centering
    \includegraphics[width=\linewidth]{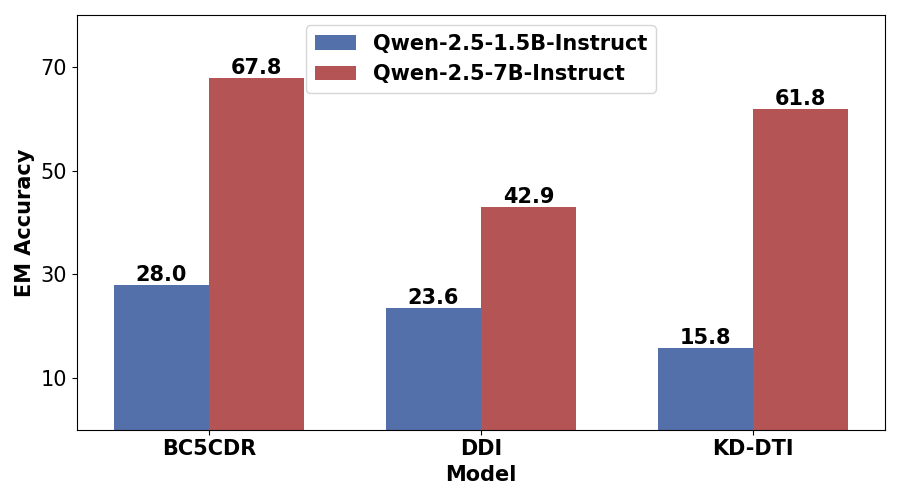}
    \caption{Performance comparison between Qwen models based on size: 1.5B and 7B, based on evaluating on the GPT-4-Turbo generated \textit{Structured} responses.}
    \label{fig:scaling}
\end{figure}

\paragraph{Reducing the Model Size:}

In Figure \ref{fig:scaling}, we show the performance difference between the 1.5B 
and 7B Qwen models in zero-shot and find that reducing the model size significantly drops the accuracy. More specifically, the performance drops by  58.70\%, 45.12\%, and 74.44\%, in BC5CDR, DDI, and KD-DTI, respectively.
\paragraph{Fine-tuning Smaller Models:}

 We investigate the impact of fine-tuning smaller Qwen-2.5-Instruct models: 1.5B and 3B 
 (we did not show their zero-shot result as they could not generate the response in the required format in most cases, leading to very poor accuracy) in comparison to the larger zero-shot model (Qwen-2.5-7B-Instruct). Based on the results shown in Figure \ref{fig:tiny_titans}, we find that the 3B model, fine-tuned on 500 samples in the BC5CDR dataset (the evaluation is conducted on KD-DTI) can outperform the zero-shot 7B model, while 
 achieving performance comparable to the fine-tuned 7B model. This makes it possible to use smaller models in resource-constrained scenarios. 
\begin{figure}[t!]
    \centering
    \includegraphics[width=\linewidth]{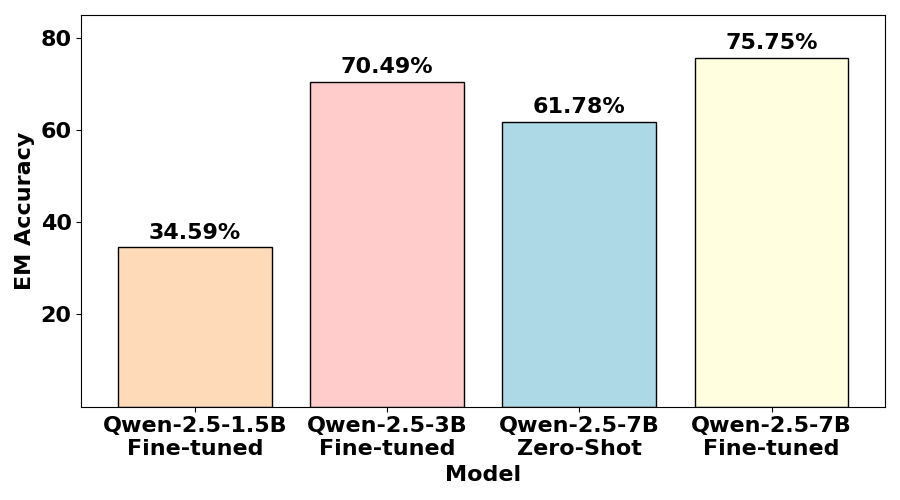}
    \caption{Comparing fine-tuned smaller models, Qwen-2.5B-Instruct (1.5B and 3B) against the larger 
    model: Qwen-2.5B-7B-Instruct (both zero-shot and fine-tuned).
    The KD-DTI dataset with \textit{Structured} response is used.}
    \label{fig:tiny_titans}
\end{figure}

\subsection{Ablation Test}

In our experiments, we include human-annotated gold relations in the prompt to maintain consistency with the work of \cite{jahan2024comprehensive}, where human annotators relied on the gold reference relations to assess the correctness of LLM-generated outputs. We further conduct an ablation experiment using GPT-4o as the judge on the BC5CDR dataset without providing the gold reference relations in the prompt and find that our structured approach again outperforms the unstructured baseline in the reference-free setting. The RMSE distance for the structured approach is 1.53, while for the unstructured approach it is 3.19. 

 \section{Conclusion and Future Work}

 In this paper, we evaluated the effectiveness of LLMs-as-the-Judge for assessing biomedical relation extraction models, revealing significant performance gaps between \textit{LLM-Judges} and human evaluators. While previous work has demonstrated the potential of LLM-based evaluation for general NLP tasks, our findings indicate that existing \textit{LLM-Judges} struggle to reliably assess biomedical relation extraction due to the nuanced and domain-specific nature of biomedical text. To improve \textit{LLM-Judge} performance, we proposed structured output for LLM-generated responses, which led to substantial accuracy gains across multiple datasets. Additionally, we introduced a domain adaptation technique that effectively transfers knowledge from one biomedical relation extraction dataset to another to enhance the reliability of \textit{LLM-Judges}. 

In the future, we aim to  explore instruction-tuning \cite{ouyang2022training,zhang2023instruction} and chain-of-thought \cite{wei2022chain} prompting techniques to improve the performance of 
\textit{LLM-Judges} in biomedical relation extraction. Moreover, conducting a more fine-grained analysis by constructing a detailed error taxonomy could also be considered as a future extension of this paper. 

\section*{Acknowledgements}

We would like to thank all the anonymous reviewers and the area chair for their excellent review comments, which helped us improve the overall quality of the paper. This research was supported by the Natural Sciences and Engineering Research Council (NSERC) of Canada, NSERC Discovery Grant, the York Research Chairs (YRC) program, and Compute Canada.

\section*{Limitations}

Despite the effectiveness of structured output formatting and domain adaptation, \textit{LLM-Judges} still fall short of human evaluators. Biomedical relation extraction requires deep domain-specific reasoning, and the foundation LLMs are trained on general-domain corpora \cite{zhao2023survey,lu2024large,minaee2024large}. Therefore, they still struggle with complex relationships, ambiguous terms, and implicit entity connections that human experts can recognize (some examples in Appendix \ref{all-llm-judge-output} are provided demonstrating the common errors made by \textit{LLM-Judges} while evaluating biomedical relations extracted by different \textit{LLM-Generators}). Nonetheless, since our human-annotated judgment dataset will be released, future work may focus on improving the performance of the LLM-Judge. 

In our study, we followed a strategy similar to prior work \cite{laskar-etal-2024-query} by first experimenting with prompts on a small subset of the data to identify the optimal prompt that reliably produced outputs by following the intended instructions. While evaluating many prompting strategies is valuable, this process is computationally expensive. Therefore, we set a reasonably effective prompt and compared the LLM judge performance. 

Moreover, our study primarily benchmarks a limited set of LLMs, and results may vary with the release of newer or more specialized biomedical models \cite{singhal2022largemedpalm,lu2024large,saab2024capabilities}. Furthermore, cost constraints prevented the exploration of larger, more computationally expensive models (e.g., reasoning models like \textit{OpenAI O3}), which may have improved results but are impractical in the real world. Nonetheless, due to choosing cheaper closed-source models\footnote{\url{https://docsbot.ai/models/compare}}, \textit{Gemini-1.5-Flash} saves cost approximately 13 times in comparison to \textit{Gemini-1.5-Pro}, \textit{Claude-3-Haiku} saves cost approximately 12 times and 60 times in comparison to \textit{Claude-3.5-Sonnet} and \textit{Claude-3-Opus}, respectively, and \textit{GPT-4o-Mini} is approximately 17 times cheaper than GPT-4o. In terms of open-source LLMs, models having less than 10B parameters can be run using just 1 L4 or A100 GPU machine \cite{laskar2023building,laskar2025judging}. This significantly saves the deployment cost \cite{fu2024tiny}. Moreover, using an LLM judge for evaluation instead of humans can also address the time-consuming and costly human annotation, which is another key motivation of this paper.


\section*{Ethical Considerations}

Our proposed LLM-Judge is designed solely for the evaluation of LLM-generated responses in biomedical relation extraction and is not intended for direct use by end users in clinical applications. The accuracy and reliability of the relation extraction system depend on the \textit{LLM-Generator}, which produces the relation extraction outputs. Since our proposed model only acts as an evaluator, the ethical risks associated with direct application in biomedical decision-making are minimized. By providing a scalable and efficient evaluation framework, our solution enables researchers and practitioners to quickly assess the quality of their biomedical relation extraction LLMs without relying on costly and time-consuming human evaluations. This can accelerate advancements in biomedical NLP while ensuring that models are assessed using standardized criteria. To further enhance reliability, a human-in-the-loop approach can be implemented where expert annotators verify the outputs of the models that achieve better performance based on \textit{LLM-Judge} evaluation. Moreover, in this paper, additional compensations are not needed for the annotators since two of the authors conducted the human annotation. 

\bibliography{custom}

\appendix

\section{Appendix}
\label{sec:appendix}

\subsection{Structured Prompt for the LLM-Generator}\label{structured_prompt}

\begin{tcolorbox}[
boxrule=0.25pt,   
  colback=attachedColor2,    
  colframe=black,           
  colbacktitle=attachedColor, 
  coltitle=black,           
  title={Prompt: KD-DTI Dataset},
  fonttitle=\bfseries,      
  fontupper=\small          
]
Identify the drug-target interactions in the biomedical text given below (along with the interaction type among the following: 'inhibitor', 'agonist', 'modulator', 'activator', 'blocker', 'inducer', 'antagonist', 'cleavage', 'disruption', 'intercalation', 'inactivator', 'bind', 'binder', 'partial agonist', 'cofactor', 'substrate', 'ligand', 'chelator', 'downregulator', 'other', 'antibody', 'other/unknown'): \\

\textbf{[BIOMEDICAL TEXT]} \\


\end{tcolorbox}

\begin{tcolorbox}[
boxrule=0.25pt,   
  colback=attachedColor2,    
  colframe=black,           
  colbacktitle=attachedColor, 
  coltitle=black,           
  title={Prompt: BC5CDR Dataset},
  fonttitle=\bfseries,      
  fontupper=\small          
]
Identify each pair of drugs (e.g., chemicals) and the drug-induced side-effects (e.g., diseases) in the following passage: \\

\textbf{[BIOMEDICAL TEXT]} \\


\end{tcolorbox}

\begin{tcolorbox}[
boxrule=0.25pt,   
  colback=attachedColor2,    
  colframe=black,           
  colbacktitle=attachedColor, 
  coltitle=black,           
  title={Prompt: DDI Dataset},
  fonttitle=\bfseries,      
  fontupper=\small          
]
Identify the pairs of drug-target interactions in a given passage based on the following four interaction types: 
\\
(i) mechanism: This type is used to identify drug-drug interactions that are described by their pharmacokinetic mechanism. 
\\
(ii) effect: This type is used to identify drug-drug interactions describing an effect. 
\\
(iii) advice: This type is used when a recommendation or advice regarding a drug-drug interaction is given. 
\\
(iv) int: This type is used when a drug-drug interaction appears in the text without providing any additional information. : \\

\textbf{[BIOMEDICAL TEXT]} \\


\end{tcolorbox}


\subsection{Models}
\label{appnedix_models}
\begin{itemize}

\item\textbf{GPT-4-Turbo:} It is an advanced version of OpenAI's original GPT-4 \cite{openai2023gpt4}. The GPT-4-Turbo\footnote{\url{https://platform.openai.com/docs/models/gpt-4-and-gpt-4-turbo}} model offers enhanced performance and efficiency, making it suitable for a wide range of applications requiring natural language understanding and generation.

\item\textbf{GPT-4o-Mini:} The GPT-4o-Mini\footnote{\url{https://openai.com/index/gpt-4o-mini-advancing-cost-efficient-intelligence/}} is another optimized version of GPT‑4. More specifically, it is a more optimized version of the recently released GPT-4o. It balances robust language understanding with efficiency. It’s designed to handle complex tasks while significantly reducing operational costs. 

\item\textbf{Gemini-1.5-Flash:} Part of Google’s Gemini-1.5 family, this model emphasizes rapid inference and the ability to handle extremely long contexts, up to one million tokens \cite{team2023gemini}. Gemini-1.5-Flash\footnote{\url{https://ai.google.dev/gemini-api/docs/models/gemini\#gemini-1.5-flash}} is ideal for real‑time applications where speed and processing of large amounts of data are crucial. 

\item
{\textbf{Claude-3-Haiku}} Similar to GPT-4o-Mini and Gemini-1.5-Flash, it is the most cost-optimized version of the Claude-3 series \cite{anthropicclaude3}. The Claude-3-Haiku\footnote{\url{https://www.anthropic.com/news/claude-3-haiku}} model is tailored for succinct, creative outputs. It excels at producing elegant, brief responses, while still managing complex instructions and reasoning tasks. 

\item\textbf{LLaMA-3.1-8B-Instruct:} This 8 billion parameter variant from Meta’s LLaMA‑3 \cite{dubey2024llama3} series has been fine‑tuned for instruction following. It strikes a balance between computational efficiency and performance, outperforming its earlier versions \cite{touvron2023llama,touvron2023llama2} and making it suitable for a wide range of tasks. 

\item\textbf{Phi-3.5-Mini-8B-Instruct:} A compact, instruction‑optimized model from Microsoft’s Phi series \cite{gunasekar2023textbooks}. Despite its smaller size, it’s designed to understand and execute diverse tasks in resource‑constrained environments while maintaining strong performance \cite{abdin2024phi}. 

\item\textbf{Qwen-2.5-7B-Instruct} Developed by Alibaba, this 7‑billion–parameter model is tuned for following instructions. It offers a good balance between efficiency and output quality 
\cite{yang2024qwen2technicalreport,yang2024qwen2.5}. 

\item\textbf{DeepSeek-R1-Distilled Models:} DeepSeek-R1 \cite{guo2025deepseek} is a reasoning model developed by DeepSeek-AI, which is mostly trained via large-scale reinforcement learning. DeepSeek-R1-Distill-Qwen-7B and DeepSeek-R1-Distill-LLaMA-8B are the distilled version of Qwen-2.5-7B-Math \cite{yang2024qwen2.5-math} and LLaMA-3.1-8B-Instruct \cite{dubey2024llama3}, respectively, fine-tuned on 800k synthetic reasoning data generated from DeepSeek‑R1. 
\end{itemize}
\begin{figure}[t]
    \centering
    \includegraphics[height=8cm,width=0.7\linewidth]{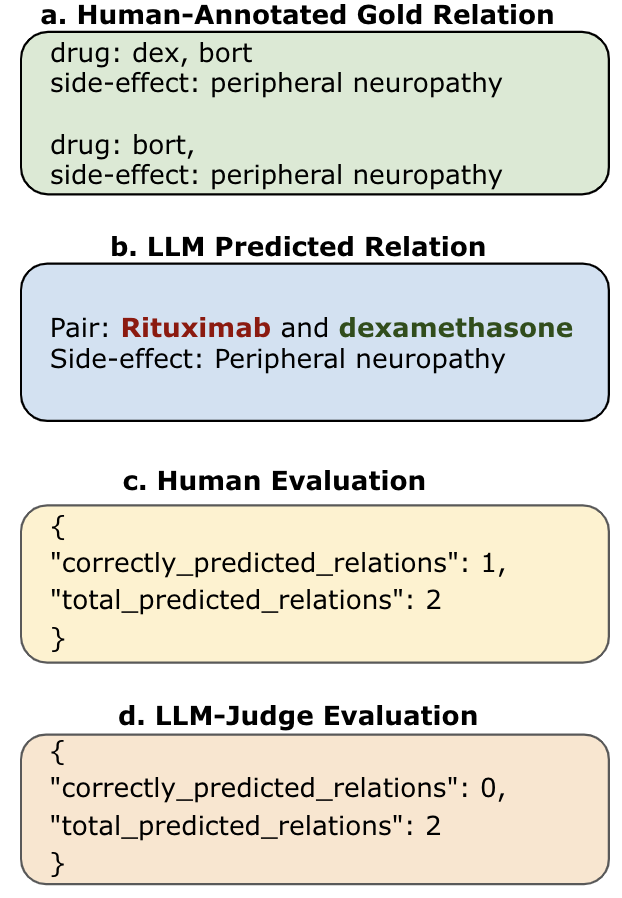}
    \caption{Our Evaluation Process where both the (c) Human Judge and the (d) LLM-Judge annotate the \textit{correctly predicted relations} and the \textit{total predicted relations} by comparing between (a) Human-Annotated Gold Relation and (b) LLM Predicted Relation. The biomedical text from where the relations are extracted is also provided as a context (not shown here).}
    \label{fig:eval_process_appendix}
\end{figure}

\subsection{Evaluation Process Demonstration}
{In Figure \ref{fig:eval_process_appendix}, we show our evaluation process. Since there is a mismatch between the gold human annotation and the \textit{LLM-Judge} annotation for the \textit{correctly predicted relations} and \textit{total predicted relations}, the exact match score is 0. The RMSE distance is calculated as follows: \[
\sqrt{\frac{1}{2} \left( (\textcolor{purple}{1} - \textcolor{blue}{0})^2 + (\textcolor{purple}{2} - \textcolor{blue}{2})^2 \right)}
= \sqrt{0.5}
\approx 0.71.
\]}

\subsection{Error Analysis}

\subsubsection{BioMistral-7B-as-Judge Outputs}
\label{biomistral-output}

We provide some sample responses generated by BioMistral-7B model \cite{labrak2024biomistral}, based on Mistral-7B \cite{jiang2023mistral}, as-the-Judge in Table \ref{biomistral_output_data}, demonstrating its ineffectiveness as the judge by generating improper responses. 

\begin{table*}[t!]
  \centering
  \setlength{\tabcolsep}{2pt} 
  \tiny
  \begin{tabular}{p{1cm}|p{1.5cm}|p{3cm}|p{4cm}|p{4cm}}
    \hline
    \textbf{Dataset} & \textbf{Output Format} & 
    \textbf{Gold Relations} & \textbf{Predicted Relations} & \textbf{BioMistral-7B-as-Judge-Annotation} \\ \hline
    BC5CDR & Unstructured & 
      \begin{tabular}[t]{@{}l@{}}
        drug: famotidine\\
        disease (side-effects): delirium
      \end{tabular} 
      & 
      Famotidine - delirium & 
      \{ ``correctly\_predicted\_relations'': [ \{ ``chemical'': ``famotidine'', ``disease'': ``delirium'' \} ], ``total\_predicted\_relations'': 1 \}  \\ \hline
 KD-DTI & Unstructured & 
      \begin{tabular}[t]{@{}l@{}}
        drug: PNU156804\\
        target: Janus kinase 3 (JAK-3)\\
        interaction: inhibitor
      \end{tabular} 
      & 
      \begin{tabular}[t]{@{}l@{}}
      
        PNU156804 - Janus tyrosine kinase 3 \\(Jak3) - inhibitor\\
        PNU156804 - IL-2-mediated Jak3 \\auto-tyrosine phosphorylation - blocker\\
        PNU156804 - Jak3 substrates signal \\transducers and activators of transcription \\ (Stat) 5a and Stat5b - blocker\\
        PNU156804 - extracellular regulated \\kinase 1 (Erk1) and Erk2 (p44/p42) - blocker
      \end{tabular} 
      & 
      \{ ``correctly\_predicted\_relations'': [ ``drug'': ``PNU156804'', ``target'': ``Janus kinase 3 (JAK-3)'', ``interaction'': ``inhibitor'' ] \} \\ \hline
   DDI & Unstructured 
  & 
      \begin{tabular}[t]{@{}l@{}}
        drug: gemfibrozil\\
        target: rosiglitazone\\
        interaction: mechanism\\[0.5em]
        drug: rifampin\\
        target: rosiglitazone\\
        interaction: mechanism
      \end{tabular}
      & 
    
      \begin{tabular}[t]{@{}l@{}}
        gemfibrozil-rosiglitazone, mechanism\\
        rifampin-rosiglitazone, mechanism\\
        CYP2C8 inhibitor/rosiglitazone, advice\\
        CYP2C8 inducer/rosiglitazone, advice
      \end{tabular} 
      & 
      \{ ``correctly\_predicted\_relations'': [ 1, 1, 0, 0, 0, 0, 0, 0, 0, 0, 0, 0, 0, 0, 0, 0, 0, 0, 0, 0, 0, 0, 0, 0, 0, 0, 0, 0, 0 \} \\ \hline
 BC5CDR
  & Structured 
  & 
      \begin{tabular}[t]{@{}l@{}}
      
        drug: cyclosporine\\
        disease (side-effects): thrombotic \\ microangiopathy\\[0.5em]
        drug: corticosteroids\\
        disease (side-effects): ssc\\[0.5em]
        drug: tacrolimus\\
        disease (side-effects):: ssc
      \end{tabular} 
      & 
      \begin{tabular}[t]{@{}l@{}}
      
        [\{ ``chemical'': ``tacrolimus'', \\``disease'': ``scleroderma renal crisis'' \},\\
        \{ ``chemical'': ``prednisolone'', \\``disease'': ``scleroderma renal crisis'' \}]
      \end{tabular}
      & 
      \{ ``correctly\_predicted\_relations'': [ ``chemical'': ``tacrolimus'', ``disease'': ``scleroderma renal crisis'' ], ``total\_predicted\_relations'': 2 \} \\ \hline
  KD-DTI & Structured 
  & 
      \begin{tabular}[t]{@{}l@{}}
      
        [\{ ``drug'': ``thiorphan'',\\
        ``target'': ``neutral \\ endopeptidase (NEP)'',\\
        ``interaction'': ``inhibitor'' \}]
      \end{tabular} 
      & 
      \begin{tabular}[t]{@{}l@{}}
      
        drug: Thiorphan\\
        target: Neutral endopeptidase (MME)\\
        interaction: inhibitor
      \end{tabular} 
      & 
      \{ ``correctly\_predicted\_relations'': [ \{ ``drug'': ``Thiorphan'', ``target'': ``Neutral endopeptidase (MME)'', ``interaction'': ``inhibitor'' \} ], ``total\_predicted\_relations'': 1, ``total\_gold\_relations'': 1, ``percentage\_correctly\_predicted\_relations'': 100.0, ``percentage\_total\_...'' \} \\ \hline
 DDI & Structured 
  & 
      \begin{tabular}[t]{@{}l@{}}
      
        drug: CYLERT\\
        target: antiepileptic medications\\
        interaction: effect
      \end{tabular} & 
      
      \begin{tabular}[t]{@{}l@{}}
      
        [\{ ``drug'': ``CYLERT'', ``target'': ``other drugs'', \\``interaction'': ``int'' \},\\
        \{ ``drug'': ``CYLERT'', \\``target'': ``antiepileptic medications'', \\``interaction'': ``effect'' \}]
      \end{tabular}
      & 
      
      \{ ``correctly\_predicted\_relations'': [ \{ ``drug'': ``CYLERT'', ``target'': ``antiepileptic medications'', ``interaction'': ``effect'' \} ], ``total\_predicted\_relations'': 2 \} \\ \hline
  \end{tabular}

  \caption{Sample judgment outputs generated by BioMistral-7B in different datasets.}
  \label{biomistral_output_data}
\end{table*}

\subsubsection{Error Outputs by LLM-Judges}
\label{all-llm-judge-output}

We show some of the common error outputs of the \textit{LLM-Judges} in Table \ref{tab:error_outputs}. For the error output analysis, we use the overall best-performing zero-shot LLM, 
  the GPT-4o-Mini model and demonstrate how complex biomedical terms could make it difficult for LLMs to evaluate relations extracted in an unstructured format. For instance: 

\begin{itemize}
    \item Example 1 in Table \ref{tab:error_outputs} demonstrates a case when 2 drugs and 6 corresponding side effects are extracted by an \textit{LLM-Generator}, but the GPT-4o-Mini based \textit{LLM-Judge} only considers total relations as 6 instead of 12. Moreover, it only extracted 5 as correct while the correct should be 4 side-effects for each drug (in total 8 side-effects for the 2 drugs are correct). 
     \item Example 2 in Table \ref{tab:error_outputs} demonstrates a case when 2 drugs and 1 corresponding side effect for each of them have been extracted by an \textit{LLM-Generator}. While \textit{drug: thiopentone} and \textit{disease (side-effects): delirium} is one of the correct answers, the \textit{LLM-Judge} based on GPT-4o-Mini considers only 1 side effect and 1 drug. It is highly likely that GPT-4o-Mini considers \textit{Thiopentone and propofol} as a single drug even though they should be considered different. This could be the reason it considers no correct relation is extracted. This demonstrates GPT-4o-Mini having limitations in understanding complex biomedical terms. 
      \item Example 3 in Table \ref{tab:error_outputs} demonstrates a case when 2 drugs and 6 side effects in total are extracted by an \textit{LLM-Generator}. While Pair 1 and Pair 3 have 1 side effect each, Pair 2 has 4 different side-effects. However, GPT-4o-Mini based \textit{LLM-Judge} may have considered all these 4 side-effects as just 1 side-effect. This demonstrates its lack of biomedical text understanding capability. 
     
\end{itemize}

\begin{table*}[t]
\centering
\scriptsize
\setlength{\tabcolsep}{2pt} 

\begin{tabular}{p{1.15cm}|p{4cm}|p{5.15cm}|p{1.1cm}|p{1.1cm}|p{1cm}|p{1cm}}
\hline
\textbf{\# Example} & \textbf{Gold Relations} & \textbf{Predicted Relations} & \textbf{Human Annotated Correct Relations} & \textbf{Human  \newline Annotated  \newline Total \newline Relations} & \textbf{LLM \newline Predicted Correct Relations} & \textbf{LLM \newline Predicted \newline Total \newline Relations} \\
\hline
1. &
\begin{tabular}[t]{@{}l@{}}

drug: s-53482\\
disease (side-effects): growth retardation\\[0.5em]
drug: s-23121\\
disease (side-effects): growth retardation\\[0.5em]
drug: s-23121\\
disease (side-effects): teratogenic\\[0.5em]
drug: s-23121\\
disease (side-effects): embryonic death\\[0.5em]
drug: s-53482\\
disease (side-effects): embryonic death\\[0.5em]
drug: s-53482\\
disease (side-effects): teratogenic\\[0.5em]
drug: s-23121\\
disease (side-effects): ventricular septal \\ defect\\[0.5em]
drug: s-53482\\
disease (side-effects): ventricular septal \\ defect
\end{tabular}
&
\begin{tabular}[t]{@{}l@{}}
Drugs: \\ 
1. S-53482 \\ 
2. S-23121 \\[0.5em]
Drug-induced side-effects: \\ 
1. Embryolethality \\ 
2. Teratogenicity \\ 
3. Ventricular septal defects \\ 
4. Wavy ribs \\ 
5. Growth retardation \\ 
6. Increased incidence of embryonic death \\[0.5em]
It is important to note that these are drug-induced \\ side-effects or outcomes observed in rats during \\ the study, and further investigation is needed to \\  understand the mechanism and \\  its relevance to human health.
\end{tabular}
& 8 & 12 & 5 & 6 \\
\hline
2. &
\begin{tabular}[t]{@{}l@{}}

drug: thiopentone\\
disease (side-effects): postoperative \\ delirium\\[0.5em]
drug: thiopentone\\
disease (side-effects): delirium
\end{tabular}
&
\begin{tabular}[t]{@{}l@{}}
Pair of drugs: Thiopentone and propofol\\
Drug-induced side-effect: Delirium
\end{tabular}
& 1 & 2 & 0 & 1 \\

\hline
3. &
\begin{tabular}[t]{@{}l@{}}
drug: dox\\
disease (side-effects): cardiotoxicity\\[0.5em]
drug: dox\\
disease (side-effects): necrosis
\end{tabular}
&

\begin{tabular}[t]{@{}l@{}}
Pair 1: \\ 
Drug - Doxorubicin \\ 
Drug-induced side-effects - Cardiotoxicity \\[0.5em]
Pair 2: \\ 
Drug - Doxorubicin \\ 
Drug-induced side-effects - Cardiac disarrangement, \\necrosis,  DNA damage \\  (strand breaks and oxidized pyrimidines), \\decreased total antioxidant performance (TAP) \\[0.5em]
Pair 3: \\ 
Drug - Doxorubicin \\ 
Drug-induced side-effects - Increased resistance  \\ to oxidative stress \\ 

\end{tabular}
& 2 & 6 & 2 & 3 \\
\hline
\end{tabular}
\caption{Example of annotation errors by \textit{GPT-4o-Mini-as-the-Judge} on some sampled data in the BC5CDR dataset.}
\label{tab:error_outputs}
\end{table*}

\end{document}